\documentclass{article} 
\usepackage{iclr2027_conference,times}
\usepackage[T1]{fontenc}
\usepackage{microtype}
\usepackage{booktabs}
\usepackage{graphicx}
\definecolor{readrow}{RGB}{226,239,248}
\newcommand{\readcell}[1]{\begingroup\setlength{\fboxsep}{1pt}\colorbox{readrow}{\strut #1}\endgroup}
\newsavebox{\fittabbox}
\newcommand{\fittable}[1]{%
  \sbox{\fittabbox}{#1}%
  \ifdim\wd\fittabbox>\linewidth
    \resizebox{\linewidth}{!}{\usebox{\fittabbox}}%
  \else
    \usebox{\fittabbox}%
  \fi}
\usepackage{amsmath}
\usepackage{amssymb}
\usepackage{array}
\usepackage[hidelinks]{hyperref}
\usepackage{caption}
\usepackage{placeins}
\usepackage{algorithm}
\usepackage{algpseudocode}

\title{New LoRA Skills Should Read but Never Write}
\author{
\textbf{Zeyan Li\textsuperscript{1} \quad
Panqi Yang\textsuperscript{2} \quad
Qirong Guo\textsuperscript{3} \quad
Shengda Zhuo\textsuperscript{4} \quad
Siyuan Qiu\textsuperscript{1} \quad
}\\
\textbf{
Hu Xu\textsuperscript{1} \quad
Chun Li\textsuperscript{1} \quad
Jianfeng Xu\textsuperscript{1}
}\\[4pt]
\textsuperscript{1}Shanghai Jiao Tong University \\
\textsuperscript{2}Xi'an Jiaotong University \\
\textsuperscript{3}The Hong Kong University of Science and Technology (Guangzhou) \\
\textsuperscript{4}Jinan University
}

\iclrfinalcopy

\begin{document}
\maketitle

\begin{abstract}
Low-rank adapters (LoRA) make it cheap to fine-tune a large language model once per task, but combining several independently trained adapters into one model remains difficult: merging the updates in weight space causes interference, retraining on all task data is expensive, and routing between separate adapters gives up the goal of a single combined model. We trace the difficulty to two choices that every composition method makes implicitly. A LoRA update admits infinitely many equivalent factorizations; the choice among them is invisible while an adapter serves alone, but it determines what a learned interaction between adapters can see. A coupling between an old skill and a new one can likewise point in either direction, and the direction decides whether the old skills keep computing what they computed before. We introduce READ (Read-only Expansion of Adapter Deltas), which fixes both choices: each adapter is rewritten into a balanced canonical form that preserves its update exactly, and the coupling grows in one direction only, so a new skill can read the input subspaces of old skills but cannot write into their output subspaces. The only trainable object at each append is the new skill's row of the coupling matrix, and the composed update folds into the base weights with no inference cost, routing, or task-specific rules. We evaluate READ across four benchmark suites and two model families, adding skills one at a time. Across several families, READ improves every suite average over the strongest published baselines built from the same adapters---by more than twenty points on SuperGLUE and more than seven points on the domain suite---and nearly all complete addition sequences end above every direct baseline. Factor coordinates and coupling direction, which a lone adapter never exposes, are what decide whether composed skills survive.
\end{abstract}

\section{Introduction}
\label{sec:intro}

Fine-tuning large language models with LoRA adapters is standard practice, and each task yields a small, reusable module \citep{hu2022lora,houlsby2019adapters}. A practical system therefore accumulates many independently trained adapters, but combining them into one model remains difficult. Adding updates in weight space causes interference \citep{ilharco2023task,wortsman2022soups,matena2022fisher}. Joint retraining on all task data is expensive and can damage the skills already in the model. Routing keeps every adapter separate \citep{pfeiffer2021adapterfusion}, but then the system never becomes a single combined model at all.

Existing methods either improve merge arithmetic or avoid merging. TIES, DARE, LoRA--LEGO, and KnOTS repair, decompose, or align updates before averaging \citep{yadav2023ties,yu2024dare,zhao2024loralego,stoica2024knots,entezari2022permutation,ainsworth2023gitrebasin}, but a new adapter still requires rebuilding the merge and moving old skills again. LoRAHub, PHATGOOSE, LoRA-Mixer, and GraftLLM keep modules separate through coefficients, experts, routers, or grafts \citep{huang2024lorahub,muqeeth2024phatgoose,li2026loramixer,du2026graftllm}. The old skills survive, but the serving system keeps an extra object and never learns a folded interaction among skills.

\begin{figure}[t]
\centering
\includegraphics[width=\linewidth]{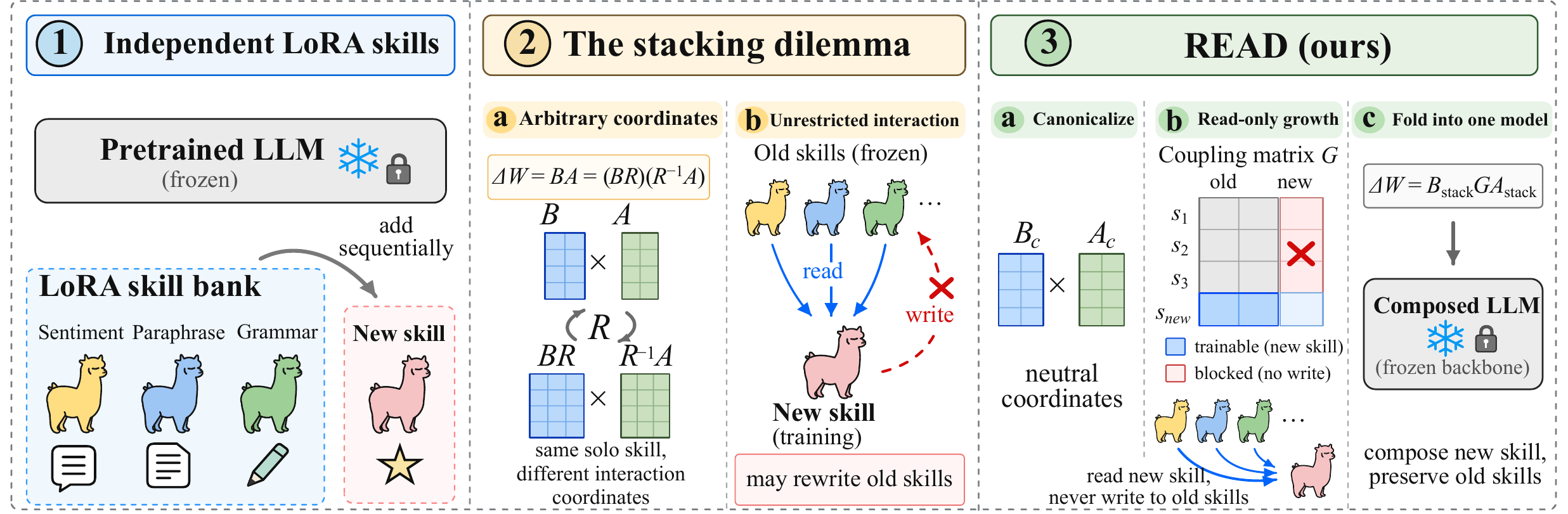}
\caption{Overview. Naive composition of independently trained LoRA skills leaves two freedoms unmanaged, arbitrary factor coordinates and unrestricted write directions. READ canonicalizes the factors, trains only the new skill's read row in $G$, blocks writes into old skills, and folds $B_{\mathrm{stack}}GA_{\mathrm{stack}}$ into one update.}
\label{fig:read}
\end{figure}

These approaches treat an adapter as a finished object whose internal form does not matter. Composition is where this form starts to matter, because a LoRA update $\Delta W = BA$ can also be written as $(BR)(R^{-1}A)$ for any invertible $R$. This gauge freedom is invisible while an adapter serves alone, but a learned coupling between adapters can inherit arbitrary factor choices rather than skill relations. Interaction also adds a directional choice, since a link between an old skill and a new one can point either way, and existing practice does not decide the direction. Composition therefore forces a question that no existing method makes explicit. When a new skill must share a model with old ones, one side has to adapt, and the choice of which side determines whether the old skills survive. Figure~\ref{fig:read} (middle) illustrates both.

\textbf{READ} (Read-only Expansion of Adapter Deltas) fixes both choices (Figure~\ref{fig:read}, right; \S\ref{sec:method}). The trained adapters form a frozen bank, joined by a small coupling matrix $G$ between their stacked factors, $\Delta W = B_{\mathrm{stack}} G A_{\mathrm{stack}}$. READ rewrites every adapter into a balanced canonical form that preserves its update exactly, and at each append trains only the new skill's read row of $G$ into the frozen bank. Everything previously learned is preserved by construction because all factors and established couplings stay frozen. The composed update folds into a single rank-$kr$ weight change, so serving costs nothing beyond the base model and needs neither a router nor a per-task head.

We evaluate READ on the GLUE, SuperGLUE, Domain, and BBH benchmarks with Llama-3.2-3B and Qwen3-4B, adding one skill at a time. On Llama-3.2-3B, READ wins the GLUE, SuperGLUE, and Domain terminal suite averages against the strongest foldable alternative, including $0.783$ versus $0.605$ on SuperGLUE and $0.887$ versus $0.846$ on Domain. On Qwen3-4B it likewise wins those three suites (GLUE $0.838$ versus $0.775$ for the strongest alternative). Across the shared 32-lineage terminal-bank comparison, READ wins 24 lineages and loses eight, all on BBH. This paper makes three contributions.
\begin{itemize}
\item We identify two choices that every LoRA composition method makes implicitly, factor coordinates and coupling direction, and show that both matter once skills interact.
\item We propose READ, a composition operator that canonicalizes adapter factors, trains only a new skill's read direction into a frozen bank, and folds the result into a single weight change that serves at base latency with no router or per-task head.
\item We evaluate READ across two model families and four benchmark suites. Across all 32 lineages the composed model beats the strongest of fourteen foldable alternatives per lineage by $+0.073$ on average (95\% CI $+0.047$ to $+0.101$), and all eight losses are confined to BBH.
\end{itemize}

\section{Related Work}
\label{sec:related}

\subsection{Merging in weight space}
Weight-space merging rests on an empirical fact. Independently fine-tuned solutions can often be averaged without destroying their task behavior \citep{wortsman2022soups,matena2022fisher}, and fine-tuned networks sit in connected low-loss regions \citep{garipov2018loss,frankle2020linear}. Task arithmetic makes this view explicit by treating each fine-tuning run as an update vector that can be added or subtracted \citep{ilharco2023task}. Later work makes this vector arithmetic less fragile, and each method repairs a different failure. TIES trims small entries and resolves sign conflicts between updates \citep{yadav2023ties}. DARE drops most entries and rescales the survivors \citep{yu2024dare}. RegMean replaces averaging with a per-neuron least-squares problem \citep{jin2023dataless}, and EMR-Merging keeps a compact shared memory so that further merges stay cheap \citep{huang2024emr}. Model Breadcrumbs removes the smallest and largest entries of each update as noise \citep{davari2024breadcrumbs}, and Model Stock pulls the merge toward the center of the fine-tuned weights \citep{jang2024modelstock}. \citet{yadav2024scale} test whether these rules still hold as models grow. A separate geometric line aligns hidden units before interpolation. Permutation symmetries explain why separately trained networks can be connected at all \citep{entezari2022permutation}, Git Re-Basin searches for a suitable alignment explicitly \citep{ainsworth2023gitrebasin}, and ZipIt extends the idea to models trained on different tasks \citep{stoica2024zipit}.

These methods produce a merged model from a completed collection of updates, and none of them learns an interaction while skills are being added. A newly arriving adapter triggers another global merge, which revises the representation of every earlier skill. READ instead asks for an append operator whose new degrees of freedom adapt to the existing bank while the bank itself stays fixed.

\subsection{Combining modules at run time}
Parameter-efficient fine-tuning makes keeping separate skills practical. The trainable object can be an inserted adapter layer \citep{houlsby2019adapters}, a learned continuous prompt \citep{li2021prefix,lester2021prompt}, only the bias terms \citep{benzaken2022bitfit}, or a pair of low-rank factors \citep{hu2022lora}. Later variants rescale activations \citep{liu2022ia3} or compress the adapter further \citep{mahabadi2021compacter}, and \citet{he2022unified} place most of these designs in one framework. Composition in this setting keeps the modules intact and chooses among them at inference. AdapterFusion learns attention over frozen adapters \citep{pfeiffer2021adapterfusion}. AdapterDrop shows that many adapters can be skipped to save computation \citep{ruckle2021adapterdrop}. LoRAHub estimates a fixed coefficient vector over adapters from a few examples \citep{huang2024lorahub}. LoRA-Mixer trains a token-level router over LoRA experts, and the router keeps running at every layer during serving \citep{li2026loramixer}. GraftLLM attaches skill packs grafted from heterogeneous source models to a lightweight host \citep{du2026graftllm}. Continual learning and sparse experts make the same trade-off from the other side. EWC and VCL protect old tasks by constraining parameter drift \citep{kirkpatrick2017ewc,nguyen2018vcl}, progressive prompts and O-LoRA allocate a new prompt or module to each task \citep{razdaibiedina2023progressive,wang2023olora}, and mixture-of-experts models route each input to a parameter subset \citep{shazeer2017moe,fedus2022switch}.

All of these leave an extra mechanism on the serving path, whether a coefficient, a switch, or an attached module. Weight-space merging removes that mechanism but treats the adapters as finished updates. READ works inside the LoRA factorization instead. It rewrites each adapter into canonical coordinates without changing its update, trains only the new skill's read direction into a frozen bank, and folds the result into the base weights, so serving uses one ordinary model with no router and no per-task head. Routing remains the appropriate choice where each solo adapter must stay intact per task, and \S\ref{sec:conclusion} returns to this boundary.

\section{The READ Operator}
\label{sec:method}

READ adds one LoRA skill to an existing bank without changing what the bank already does. A LoRA adapter stores a skill as two small matrices whose product is the weight update. When several skills share one model, two choices that a lone adapter never exposes start to matter. The first is the factor coordinates of each update, and the second is the direction of each cross-skill link. READ fixes the coordinates once, lets the new skill read the frozen bank but never write into it, and folds the result into one model.

\begin{figure}[t]
\centering
\includegraphics[width=\linewidth]{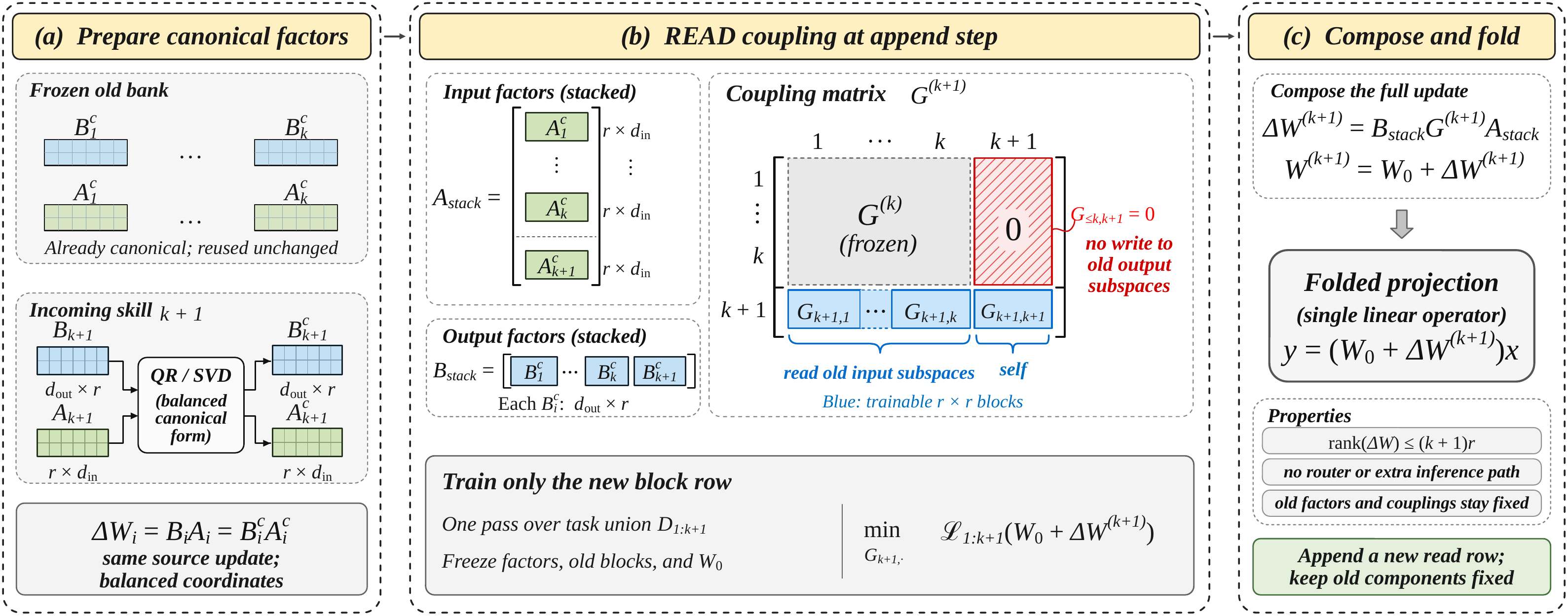}
\caption{The READ operator at one append. (a)~Every source adapter is rewritten into balanced canonical factors with $B_i^cA_i^c=B_iA_i$, so skills meet in coordinates determined by their updates. (b)~The stacked factors are joined by a coupling matrix. The old block $G^{(k)}$ stays frozen, the write column $G_{\le k,k+1}$ is fixed to zero, and only the new skill's row is trained. (c)~The product folds into the base weights $W_0$, giving one merged projection, no router, and no extra inference path.}
\label{fig:method}
\end{figure}

\subsection{The frozen skill bank}
\label{sec:bank}
Consider one projection module and omit its index. Each source skill is an independently trained generative-classification LoRA ($r{=}8$) on the $q$- and $v$-projections of every attention block, following a fixed prompt/completion recipe (\S\ref{sec:setup}). All tasks share the causal-LM head; labels are emitted as text. For $k$ skills, stack the frozen factors.
\begin{equation}
\Delta W \;=\; B_{\mathrm{stack}}\; G\; A_{\mathrm{stack}},\qquad A_{\mathrm{stack}} \in \mathbb{R}^{kr \times d_{\mathrm{in}}},\; B_{\mathrm{stack}} \in \mathbb{R}^{d_{\mathrm{out}} \times kr},\; G \in \mathbb{R}^{kr \times kr},
\label{eq:bank}
\end{equation}
$G$ is block-partitioned into $r{\times}r$ blocks $G_{ij}$, $i,j \le k$. $G_{ii}$ scales and mixes skill $i$'s own directions; $G_{ij}$, $i \ne j$, maps skill $j$'s input directions into skill $i$'s output directions. With $G$ equal to the identity the update is the naive sum of the adapters; every departure from that sum is learned inside $G$. Inference applies one effective rank-$kr$ update per module, the same functional form as a single LoRA of rank $kr$, and since factors and $G$ are frozen between appends, $\Delta W$ can be precomputed and folded into the frozen base projection $W_0$, leaving zero residual serving cost relative to the base model. The update requires no task identity, adapter routing, or per-task weight/head path at inference; the task instruction in the ordinary input prompt specifies the output label space.

\subsection{Canonicalizing every adapter}
\label{sec:canonical}
For $G$ to relate directions across skills, those directions must first be comparable. The factorization of a source update is gauge-ambiguous, since for any invertible $R$, $(B_iR)(R^{-1}A_i)=B_iA_i$. While an adapter serves alone, this freedom is invisible because the product is all the model sees. Composition exposes it. $G$ maps one skill's input basis into another's output basis, but neither skill's function determines those bases. They are accidents of optimization, set by initialization, data order, and seed, so a coupling learned between raw factorizations can encode those accidents instead of the skills. The same two adapters, factored differently by two training runs, would force different coupling values for the same functional relationship. We therefore fix the gauge before any coupling is trained, so the coordinates in which skills meet are determined by the update itself. We compute thin QR decompositions $B_i=Q_B R_B$ and $A_i^\top=Q_A R_A$, followed by $R_BR_A^\top=U\Sigma V^\top$, and set
\begin{equation}
B_i^c=Q_BU\Sigma^{1/2},\qquad A_i^c=\Sigma^{1/2}V^\top Q_A^\top.
\end{equation}
Then $B_i^cA_i^c=B_iA_i$ exactly, while $(B_i^c)^\top B_i^c=A_i^c(A_i^c)^\top=\Sigma$. Both factors carry the same metric, so no skill's basis is privileged in the interaction, and the remaining freedom is a rotation inside degenerate singular subspaces, which cannot change any product. The source skill is unchanged; only the coordinates in which it meets the others are.

\subsection{Read-only coupling}
\label{sec:readonly}
With coordinates fixed, the remaining choice is the direction of a link between an old skill and a new one. When skill $k{+}1$ arrives, its cross blocks could point either way. $G_{i,k+1}$ with $i \le k$ writes the new skill's input directions into old output bases, while $G_{k+1,i}$ lets the new skill read old input directions into its own output basis. The two directions differ in what they can damage. A nonzero $G_{i,k+1}$ adds a term to what every old skill computes on every input; old behavior is then protected only insofar as optimization happens to move it little. A nonzero $G_{k+1,i}$ adds nothing to any old output, so the old skills' function is preserved by construction. The asymmetry extends to gradients. During the append pass over the union of old and new task data (\S\ref{sec:setup}), a write block receives gradients from old-task examples, while a read block is optimized only through the new skill's output. READ therefore fixes $G_{\le k,k+1}=0$ and trains only the new skill's row and diagonal. At the append, the trainable addition is
\begin{equation}
B_{k+1}^c\!\left(G_{k+1,k+1}A_{k+1}^c+\sum_{j\le k}G_{k+1,j}A_j^c\right).
\end{equation}
The new skill may reuse old input subspaces, but it cannot write through its input directions into old output bases. All earlier blocks, every canonical factor, the backbone, and the task head are frozen.

\subsection{Appending and folding a skill}
\label{sec:append}
The incoming adapter now becomes the new bank row in five steps (Figure~\ref{fig:method}). \textbf{Extract.} Take its trained solo factors $A_{k+1}, B_{k+1}$. \textbf{Canonicalize.} Rewrite all factors into the balanced form of \S\ref{sec:canonical}; no source update changes. \textbf{Embed.} Grow $G$ from $(kr)^2$ to $((k{+}1)r)^2$ per module, install the old $G^{(k)}$ verbatim in its upper-left block, and freeze it. \textbf{Train.} Make one pass over the stage's task union, optimizing only the new row and diagonal (\S\ref{sec:setup}). \textbf{Fold.} Precompute $B_{\mathrm{stack}}GA_{\mathrm{stack}}$, add it to $W_0$, and serve with one merged projection per module.

\subsection{One READ append and its cost}
\label{sec:algorithm}

Algorithm~\ref{alg:read} makes the five steps explicit. Canonicalize and fold run per projection module; train runs jointly over all modules and is the only step that touches data.

\begin{algorithm}[t]
\caption{One READ append. Skill $k{+}1$ joins a bank of $k$ skills.}
\label{alg:read}
\begin{algorithmic}[1]
\Require frozen bank $\{B_i^c, A_i^c\}_{i\le k}$ and coupling $G^{(k)}$; incoming solo factors $B_{k+1}, A_{k+1}$; base weights $W_0$; task-union data $\mathcal{D}_{1:k+1}$
\Ensure folded weights $W_0'=W_0+B_{\mathrm{stack}}G'A_{\mathrm{stack}}$, a bank of $k{+}1$ skills
\State \textbf{Canonicalize} (\S\ref{sec:canonical}). Thin QR $B_{k+1}{=}Q_BR_B$, $A_{k+1}^{\top}{=}Q_AR_A$, then $R_BR_A^{\top}{=}U\Sigma V^{\top}$;
\Statex \hspace{\algorithmicindent}\hspace{\algorithmicindent} set $B_{k+1}^c{=}Q_BU\Sigma^{1/2}$ and $A_{k+1}^c{=}\Sigma^{1/2}V^{\top}Q_A^{\top}$ \Comment{$B^cA^c{=}BA$ exactly}
\State \textbf{Embed}. Install $G^{(k)}$ frozen in the upper-left block; set the write column $G'_{\le k,k+1}{=}0$, the new row $G'_{k+1,\le k}{=}0$, and the new diagonal $G'_{k+1,k+1}{=}0.3\,I_r$
\State \textbf{Train} (\S\ref{sec:readonly}). One pass over $\mathcal{D}_{1:k+1}$ optimizing only $G'_{k+1,k+1}$ and $G'_{k+1,\le k}$ \Comment{factors, old blocks, and $W_0$ stay frozen}
\State \textbf{Fold}. $W_0' \leftarrow W_0 + B_{\mathrm{stack}}G'A_{\mathrm{stack}}$ \Comment{one merged projection, zero serving cost}
\end{algorithmic}
\end{algorithm}

The costs follow the same split. Canonicalization costs two thin QR decompositions of $d{\times}r$ matrices and one $r{\times}r$ SVD per module, $O(dr^2+r^3)$ in the projection dimension $d$, under a million flops at Llama sizes, negligible next to any training step. Embedding places the frozen block and does no arithmetic. Training optimizes $(k{+}1)r^2$ floats per module, the new diagonal and its $k$ row blocks, and nothing else. That is 384 floats per module, 21{,}504 in total, at a six-skill append. Per token, the effective update costs the same as one LoRA layer of rank $kr$ (\S\ref{sec:bank}). The fold multiplies the three factors once per module, $O(d^2kr)$ multiply-adds in closed form; after it, serving runs the plain base model. Storage beyond the $k{+}1$ adapters is the coupling itself, $(kr)^2$ floats per module, which is 129k across the 56 Llama $q/v$ projections at $k{=}6$, about 4.7\% of one adapter. The per-append trainable set grows linearly in $k$ and the coupling storage quadratically; both remain small fractions of one adapter over the measured range.

\section{Experiments}
\label{sec:experiments}

\subsection{Setup}
\label{sec:setup}

\label{sec:population-training}
We evaluate Llama-3.2-3B-Instruct \citep{grattafiori2024llama3} and Qwen3-4B \citep{yang2025qwen3} on GLUE-6 \citep{wang2019glue}, SuperGLUE-4, BBH-6, and Domain-3. Each task first trains an independent generative-classification LoRA source adapter ($r{=}8$, $\alpha{=}16$, $q/v$ projections, shared causal-LM head) under one fixed recipe. Every READ append then trains one pass over the full stage task union with the same optimizer, schedule, batch, and length for all compared growth arms; the manipulated variable is the factor coordinate system and/or the trainable mask on $G$, not data exposure or tuning. The comparison therefore isolates the foldable interaction rule, since appending itself still trains on old-task data. All final evaluations use full configured splits, batch 64, input length 512, and each task's official primary metric; suite ``terminal macro'' is the mean over task scores.

\label{sec:arms}
The headline campaign contains 32 READ lineages (2 models $\times$ 4 suites $\times$ 2 seeds $\times$ 2 task orders) and 120 adjacent appends. The terminal comparison matches each final bank against the strongest of fourteen \emph{exact-source foldable} alternatives: seven tensor merges \citep{wortsman2022soups,matena2022fisher,ilharco2023task,jin2023dataless,yadav2023ties,yu2024dare,davari2024breadcrumbs}, three direct operators \citep{zhao2024loralego,stoica2024knots,huang2024lorahub}, and four recent published systems \citep{zhang2025osrm,prabhakar2025lorasoups,he2026compressmerge,lee2026nsc}. The append comparison applies the preregistered rule, requiring at least half the Soup-to-refit incoming gap closed, mean old-task drop no worse than 2 points, and no old task worse than 5 points. A compact gauge check tests the coordinate choice, and the append ledger traces bank-level stability across additions. Intervals are bootstrap 95\% confidence intervals over lineages.

\begin{table}[!t]
\centering\small
\caption{Terminal-bank macro comparison across READ and fourteen foldable alternatives.
Cells average the four complete terminal lineages per suite--model; the first baseline row is the per-lineage strongest alternative used for the primary comparison.}
\label{tab:readfinal-matrix}
\fittable{
\begin{tabular}{@{}lrrrrrrrrrr@{}}
\toprule
 & \multicolumn{4}{c}{Llama-3.2-3B} & \multicolumn{4}{c}{Qwen3-4B} & & \\
\cmidrule(lr){2-5}\cmidrule(lr){6-9}
Method & GLUE-6 & SuperGLUE-4 & Domain-3 & BBH-6 & GLUE-6 & SuperGLUE-4 & Domain-3 & BBH-6 & Avg & $\Delta$R \\
\midrule
\readcell{\textbf{READ (ours)}} & \readcell{\textbf{0.823}} & \readcell{\textbf{0.783}} & \readcell{\textbf{0.887}} & \readcell{0.445} & \readcell{\textbf{0.838}} & \readcell{\textbf{0.797}} & \readcell{\textbf{0.852}} & \readcell{0.770} & \readcell{\textbf{0.774}} & \readcell{--} \\
\emph{Strongest alt. (per lineage)} & 0.631 & 0.605 & 0.846 & 0.459 & 0.775 & 0.709 & 0.794 & 0.793 & 0.702 & $+0.073$ \\
\midrule
Soup & 0.016 & 0.547 & 0.695 & 0.392 & 0.339 & 0.550 & 0.666 & 0.772 & 0.497 & $+0.277$ \\
Fisher & 0.098 & 0.425 & 0.640 & 0.406 & 0.363 & 0.000 & 0.365 & 0.764 & 0.383 & $+0.392$ \\
Task Arithmetic & $-0.083$ & 0.550 & 0.682 & 0.421 & 0.087 & 0.574 & 0.653 & 0.776 & 0.458 & $+0.317$ \\
RegMean & 0.631 & 0.605 & 0.844 & 0.439 & 0.775 & 0.619 & 0.794 & 0.768 & 0.684 & $+0.090$ \\
TIES & 0.047 & 0.455 & 0.617 & 0.425 & 0.154 & 0.432 & 0.403 & 0.779 & 0.414 & $+0.361$ \\
DARE & $-0.085$ & 0.551 & 0.678 & 0.429 & 0.083 & 0.573 & 0.657 & 0.781 & 0.458 & $+0.316$ \\
Breadcrumbs & $-0.063$ & 0.546 & 0.663 & 0.411 & 0.226 & 0.529 & 0.588 & 0.789 & 0.461 & $+0.313$ \\
\midrule
LoRA--LEGO & $-0.097$ & 0.543 & 0.785 & 0.392 & 0.074 & 0.488 & 0.723 & \textbf{0.790} & 0.462 & $+0.312$ \\
KnOTS--TIES & 0.238 & 0.537 & 0.704 & 0.394 & 0.342 & 0.477 & 0.625 & 0.748 & 0.508 & $+0.266$ \\
LoRAHub--5shot & 0.209 & 0.567 & 0.811 & 0.438 & 0.210 & 0.709 & 0.647 & 0.439 & 0.504 & $+0.271$ \\
\midrule
OSRM & $-0.109$ & 0.233 & 0.178 & 0.362 & $-0.167$ & 0.000 & 0.000 & 0.717 & 0.152 & $+0.622$ \\
LoRA--Soups/CAT & 0.024 & 0.547 & 0.699 & 0.396 & 0.354 & 0.548 & 0.667 & 0.759 & 0.499 & $+0.275$ \\
Compress-then-Merge & $-0.167$ & 0.366 & 0.718 & \textbf{0.459} & $-0.167$ & 0.492 & 0.624 & 0.771 & 0.387 & $+0.387$ \\
NSC & $-0.010$ & 0.546 & 0.725 & 0.394 & 0.331 & 0.565 & 0.667 & 0.755 & 0.497 & $+0.278$ \\
\bottomrule
\end{tabular}
}
\end{table}

\subsection{Q1: Does READ build a better final skill bank?}
\label{sec:canonical-terminal}

The strongest foldable alternative changes from bank to bank. RegMean is best on 18 lineages, LoRAHub--5shot on six, and Compress-then-Merge on four. Table~\ref{tab:readfinal-matrix} therefore compares READ against the best composition available from the same source adapters in each case. READ improves on that per-lineage maximum on 24 of 32 lineages, by $+0.073$ on average (95\% CI $[+0.047,+0.101]$). The eight losses are all on BBH and are narrow, $-0.014$ on Llama and $-0.023$ on Qwen at suite level.

It is worth noting who the strongest competitor usually is. RegMean, a dataless merge from 2022, occupies that row far more often than any recent system, while the newest published methods sit at the bottom of the table, with overall averages of $0.152$ for OSRM and $0.387$ for Compress-then-Merge against RegMean's $0.684$. On GLUE in particular, seven of the fourteen alternatives score below zero on Llama, which means a merged bank can end up worse than no merge on the suite's harder components. A method that has to beat the per-lineage maximum therefore has to beat a strong classical baseline, not a weak recent one.

The per-suite margins are consistent across seeds and task orders. All 24 non-BBH terminal lineages are READ wins. Llama gains most on GLUE and SuperGLUE ($+0.192$ and $+0.178$), and Qwen's margins on the three non-BBH suites are $+0.063$, $+0.088$, and $+0.058$. Domain shows the smallest Llama margin ($+0.041$), where the strongest alternative already reaches $0.846$. Read directly off Table~\ref{tab:readfinal-matrix}, READ is the column best on all seven non-BBH cells and trails the winner on the two BBH cells by only $0.014$ (Llama) and $0.020$ (Qwen).

\subsection{Q2: Can READ append skills without erasing the bank?}
\label{sec:reliability}

A strong terminal bank can hide damaging intermediate appends, so we examine every transition. An append passes when it closes at least half of the Soup-to-refit incoming gap, with mean and worst old-task changes no lower than $-2$ and $-5$ points. READ passes 72 of 92 eligible transitions across the expanded grid (Table~\ref{tab:expanded_perf}).

\begin{figure}[!t]
\centering
\includegraphics[width=0.8\linewidth]{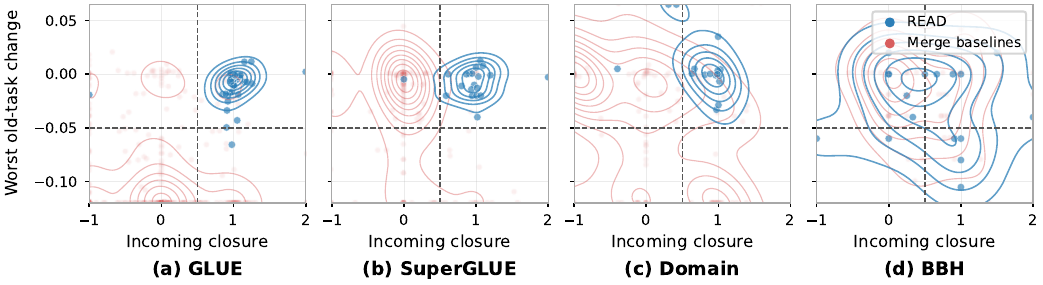}
\caption{Append phase space by suite.
Blue marks READ and red marks merge baselines; dashed lines mark pass thresholds.}
\label{fig:phase-space}
\end{figure}

The merge baselines fail on both axes of Figure~\ref{fig:phase-space} at once. Fisher's mean incoming closure is negative in six of the eight model--suite cells, down to $-4522.8\%$ on Llama GLUE, which means the merged bank ends up further from the refit target on the new task than the soup anchor it started from. Task Arithmetic, DARE, and Breadcrumbs show the same pattern at smaller magnitudes. READ's mean closure is positive in every cell, never below $57.8\%$, and its worst old-task changes stay inside the $-5$ bound wherever the append passes.

\begin{table}[t]
\centering\scriptsize
\caption{Sequential append reliability.
Pass rate and mean incoming closure are shown in separate rows; coverage gives eligible/observed transitions.}
\label{tab:expanded_perf}
\fittable{%
\begin{tabular}{@{}llcccccccc@{}}
\toprule
& & \multicolumn{2}{c}{GLUE-6} & \multicolumn{2}{c}{SuperGLUE-4} & \multicolumn{2}{c}{BBH-6} & \multicolumn{2}{c}{Domain-3} \\
\cmidrule(lr){3-4}\cmidrule(lr){5-6}\cmidrule(lr){7-8}\cmidrule(l){9-10}
Method & Metric & Llama & Qwen & Llama & Qwen & Llama & Qwen & Llama & Qwen \\
\midrule
Coverage & Eligible / observed & 19/20 & 20/20 & 11/12 & 9/12 & 7/20 & 10/20 & 8/8 & 8/8 \\
\midrule
Soup & Pass rate (\%) & 0.0 & 0.0 & 0.0 & 0.0 & 0.0 & 0.0 & 0.0 & 0.0 \\
 & Incoming closure (\%) & 0.0 & 0.0 & 0.0 & 0.0 & 0.0 & 0.0 & 0.0 & 0.0 \\
Fisher & Pass rate (\%) & 0.0 & 5.0 & 0.0 & 0.0 & 42.9 & 20.0 & 0.0 & 0.0 \\
 & Incoming closure (\%) & -4522.8 & -810.0 & -45.8 & -1481.8 & 65.5 & 17.5 & -856.5 & -594.6 \\
Task Arithmetic & Pass rate (\%) & 0.0 & 0.0 & 0.0 & 11.1 & 14.3 & 40.0 & 0.0 & 12.5 \\
 & Incoming closure (\%) & -144.0 & -57.9 & -16.6 & 24.3 & 18.2 & 49.2 & -241.8 & -85.5 \\
RegMean & Pass rate (\%) & 26.3 & 50.0 & 27.3 & 22.2 & 85.7 & 10.0 & 25.0 & 50.0 \\
 & Incoming closure (\%) & -25.9 & -276.0 & 29.1 & 25.4 & 72.3 & 72.3 & 13.8 & 89.0 \\
TIES & Pass rate (\%) & 10.5 & 0.0 & 0.0 & 22.2 & 57.1 & 30.0 & 0.0 & 0.0 \\
 & Incoming closure (\%) & 19.2 & -296.6 & 0.3 & -36.1 & 83.4 & 124.3 & -9.5 & -89.5 \\
DARE & Pass rate (\%) & 0.0 & 0.0 & 0.0 & 11.1 & 14.3 & 50.0 & 0.0 & 12.5 \\
 & Incoming closure (\%) & -157.7 & -64.3 & -14.8 & 23.5 & 19.0 & 83.3 & -239.9 & -91.7 \\
Breadcrumbs & Pass rate (\%) & 0.0 & 0.0 & 0.0 & 0.0 & 14.3 & 40.0 & 0.0 & 0.0 \\
 & Incoming closure (\%) & -180.5 & -252.6 & -25.1 & -1065.6 & 41.4 & 24.5 & -369.4 & -740.7 \\
LoRAHub & Pass rate (\%) & 15.8 & 5.0 & 9.1 & 0.0 & 42.9 & 20.0 & 0.0 & 0.0 \\
 & Incoming closure (\%) & 67.8 & -433.9 & -310.5 & -164.0 & 120.4 & 86.8 & -1014.1 & 56.0 \\
\readcell{\textbf{READ (ours)}} & \readcell{Pass rate (\%)} & \readcell{89.5} & \readcell{85.0} & \readcell{90.9} & \readcell{100.0} & \readcell{28.6} & \readcell{40.0} & \readcell{87.5} & \readcell{75.0} \\
\readcell{} & \readcell{Incoming closure (\%)} & \readcell{89.3} & \readcell{123.2} & \readcell{104.0} & \readcell{95.3} & \readcell{57.8} & \readcell{60.0} & \readcell{70.7} & \readcell{82.7} \\
\bottomrule
\end{tabular}
}
\end{table}

READ's pass rates split cleanly by suite. On GLUE, SuperGLUE, and Domain, READ passes at least $75\%$ of eligible transitions for both models. On BBH it passes only $28.6\%$ on Llama and $40.0\%$ on Qwen, and these are acquisition failures. Mean incoming closure on BBH stays positive ($57.8\%$ and $60.0\%$), so the bank still gains part of the new skill on average, but one pass over the stage union does not reliably close half the gap on these tasks. Old-bank preservation is not what fails; the mean and worst old-task bounds hold even on the failed BBH transitions. The group-level view agrees. Six of the eight model--suite groups pass the group rule. The two misses, Llama SuperGLUE and Llama Domain, fail only the retained-share clause while their mean old-task damage stays within the $-2$ bound. Read across methods, READ is also the only row of Table~\ref{tab:expanded_perf} whose incoming-closure entries are all positive.

\FloatBarrier
\subsection{Q3: Does read-only growth stabilize the skill bank?}
\label{sec:trajectory-stability}

Terminal scores do not show whether the bank followed a stable path to reach them. The gauge check quantifies the coordinate problem directly. Re-factoring the same adapters into equivalent raw factorizations shifts downstream scores by $2.8$ points on average, and canonicalization shrinks that spread to $0.7$. Without canonical coordinates, what a coupling learns depends partly on which factorization the source training happened to produce.

The append ledger then traces the bank itself (Figure~\ref{fig:pca-stability}). Merge trajectories turn and spread after individual additions, so a later skill joins a representation that has already moved away from the bank its earlier skills were composed into. READ's trajectory makes small, similarly oriented moves from the initial bank onward. The distribution panels show the same contrast in score terms. READ's closures sit mostly on the positive side of the pass threshold, and its worst old-task changes concentrate near zero, while the merge baselines spread toward retention loss.

\par\smallskip
\noindent\begin{minipage}{\linewidth}
\centering
\includegraphics[width=.90\linewidth]{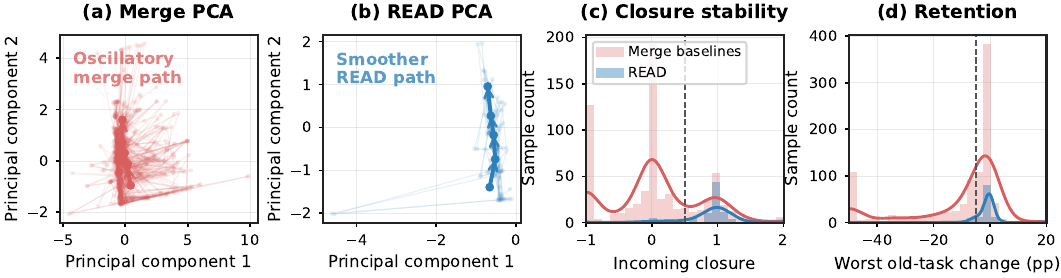}
\captionof{figure}{Trajectory and stability summaries from the append ledger.
The first two panels show PCA trajectories; the last two show closure and retention-damage distributions.}
\label{fig:pca-stability}
\end{minipage}
\par\smallskip

Two controls isolate where the stability comes from. Random norm-matched factors underperform the learned factors in 26 of 27 measured cells, so the coupling is exploiting structure in the trained updates rather than generic extra capacity. Global diagonal refits reopen established couplings and become seed sensitive, which is the failure READ's freeze rule removes by construction.

\FloatBarrier
\subsection{Q4: Does READ remain a single-model operator?}
\label{sec:deployment}

Six of the eight model--suite groups pass the whole-lineage rule, and canonicalization contracts the mean spread across equivalent source factorizations from $2.8$ to $0.7$ points. The gauge contraction is a robustness result rather than an average-utility gain: the complete component factorial places all six coordinate--direction arms within $0.017$ terminal macro, with overlapping confidence intervals. The remaining question is whether the resulting operator survives as a serving object.

\begin{figure}[t]
\centering
\includegraphics[width=\linewidth]{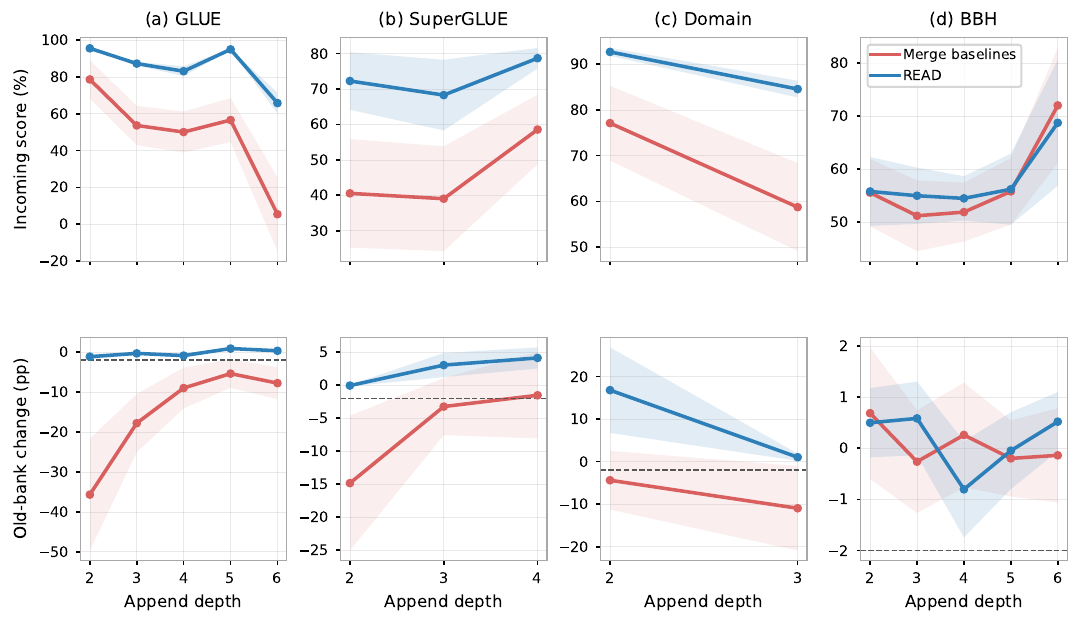}
\caption{Append dynamics by suite. Top row: incoming score at each append depth. Bottom row: mean old-bank change. Bands are one standard deviation across lineages; the dashed line marks the $-2$-point retention threshold.}
\label{fig:append-dynamics}
\end{figure}

Figure~\ref{fig:append-dynamics} first confirms that reliability is a property of the growing bank rather than of the terminal prefix alone. Old-bank change stays near zero for READ at every append depth on every suite, and incoming score follows the boundary already visible in Q2, with READ tracking the acquisition ceiling on GLUE, SuperGLUE, and Domain while BBH flattens for both READ and the merge baselines.

Folding then removes any residual READ path from that serving object. It reproduces the unfolded operator in all 128 model--configuration checks, with maximum logit deviation $4.25{\times}10^{-4}$. Before folding, the bank adds about $10$--$28\%$ mean latency across the four measured batch--length settings; after folding, the corresponding means remain within roughly one percent of the base model. The auxiliary bank occupies 79.5--166.9 MB and 6.9--17.9 million floats before folding, while the deployed checkpoint is an ordinary 6.4--12.9 GB model with no router, task identity, or extra inference branch. Against an oracle-router upper bound trained on the same adapters, the folded model averages $+0.064$ terminal macro at order-averaged bank size, so one merged projection serves at or above the routing headroom on the suites where composition adds signal, and READ pays its interaction cost while a skill is appended rather than each time the resulting bank is served.

\FloatBarrier
\section{Conclusion}
\label{sec:conclusion}

When a new skill must share a model with old ones, one side has to adapt, and the choice of which side determines whether the old skills survive. This paper traced that choice to two freedoms a lone adapter never exposes, the coordinates of its factors and the direction of each cross-skill link. Composition only works reliably when a method fixes both, instead of inheriting whatever the source training runs happened to produce.

READ fixes them. Every adapter is rewritten into balanced canonical coordinates that preserve its update exactly, and the coupling grows in one direction only, so each new skill reads the frozen bank without writing into old output subspaces. Across 32 terminal lineages this beats the strongest of fourteen foldable alternatives built from the same adapters by $+0.073$ on average, and 72 of 92 sequential appends pass a preregistered reliability rule. The composed update folds into a single rank-$kr$ weight change that serves at base-model latency and, at the same order-averaged bank size, sits above an oracle-router upper bound on the suites where composition adds signal.

The remaining boundary is BBH, where all eight terminal losses concentrate. The failures there are acquisition failures rather than retention failures. One pass over the stage union does not always give the new skill enough signal, and the read-only constraint protects the bank but cannot itself supply that missing signal. How to strengthen acquisition while keeping the bank frozen, for example by giving the new skill's row more than one pass or a richer state than the frozen factors alone, is the direction we consider most worth pursuing. Two boundaries lie outside the present evaluation and belong on that same agenda: open-ended generative composition, where the primary metric is not classification accuracy, and banks beyond six skills, where the coupling stays quadratic in $k$ even though the trained slice does not.


\clearpage
\bibliographystyle{iclr2027_conference}
\bibliography{references}

\end{document}